\documentclass{article} 
\usepackage[final]{colm2026_conference}
\usepackage{amsmath} 
\usepackage{microtype}
\usepackage{hyperref}
\usepackage{url}
\usepackage{booktabs}
\usepackage{multirow}
\usepackage{graphicx}
\usepackage{booktabs}
\usepackage{xcolor}
\usepackage{amssymb}
\usepackage{pifont}
\usepackage{enumitem}
\usepackage[table,xcdraw]{xcolor}
\usepackage{wrapfig}
\usepackage[table]{xcolor}
\usepackage{multirow}
\usepackage{booktabs}
\usepackage{makecell}
\usepackage{lineno}

\definecolor{darkblue}{rgb}{0, 0, 0.5}
\hypersetup{colorlinks=true, citecolor=darkblue, linkcolor=darkblue, urlcolor=darkblue}
\definecolor{babyblue}{HTML}{89CFF0}

\title{From Plausible to Grounded: \\Reinforcing Structured Consistency in Video MLLMs}
\author{\textbf{Yihao Quan$^1$} {\hspace{.1em}}\quad
  \textbf{Zeru Shi$^1$} {\hspace{.1em}}\quad
  \textbf{Jinman Zhao$^2$} {\hspace{.1em}}\quad
  \textbf{Ruixiang Tang$^{\dagger 1}$}{\thanks{Corresponding Author}} {\hspace{.1em}}\quad
  \vspace{.5em}\\$^1$Rutgers University $^2$University of Toronto
  \vspace{.5em}\\
  \texttt{yq207@rutgers.edu}
}
\newcommand{\methodname}{\textsc{Ours}}

\begin{document}

\ifcolmsubmission
\linenumbers
\fi

\maketitle

\begin{abstract}
Multimodal large language models (MLLMs) have achieved remarkable progress in video understanding. However, seemingly plausible outputs often suffer from poor visual and temporal grounding: a model may fabricate object existence, assign incorrect attributes, or collapse repeated events while still producing a globally reasonable caption or answer. We study this failure mode through a compositional consistency audit that decomposes a caption into supporting factual and temporal claims, investigating whether a correct high-level prediction is actually backed by valid lower-level evidence. Our top-down audit reveals that even correct root relational claims often lack reliable attribute and existence support. This indicates that standard sentence-level supervision is a weak proxy for faithful video understanding. Furthermore, when turning to reinforcement learning (RL) for better alignment, standard sentence-level rewards often prove too coarse to accurately localize specific grounding failures. To address this, we replace generic sentence-level rewards with a structured reward built from factual and temporal units. Our training objective integrates three complementary components: (1) an instance-aware scene-graph reward for factual objects, attributes, and relations; (2) a temporal reward for event ordering and repetition; and (3) a video-grounded VQA reward for hierarchical self-verification. Across temporal, general video understanding, and hallucination-oriented benchmarks, this objective yields consistent gains on open-source backbones. These results suggest that structured reward shaping is a practical route to more faithful video understanding.
\end{abstract}

\section{Introduction}
Multimodal large language models (MLLMs) ~\citep{llava, qwenvl, gpt4, gemini} have recently made strong progress in video understanding, serving as the foundation for tasks like video captioning~\citep{imagecap, msrvtt}, question answering~\citep{vqa, activitynetqa}, and agent-style interaction. Particularly in open-ended generation, the resulting text acts not just as a final prediction, but as an explicit reflection of the model's underlying understanding. Consequently, factual and temporal faithfulness is paramount. While current models can produce globally plausible outputs, they frequently fail to ground these narratives in reliable visual evidence. A minor error in object identity, attribute assignment, or event ordering can invalidate an otherwise coherent caption. This exposes a critical vulnerability: the capacity of MLLMs to generate plausible text has outpaced their ability to verify the atomic facts that support it.

\begin{wrapfigure}{r}{0.60\columnwidth}
    \vspace{-0.8em}
    \centering
    \includegraphics[width=\linewidth]{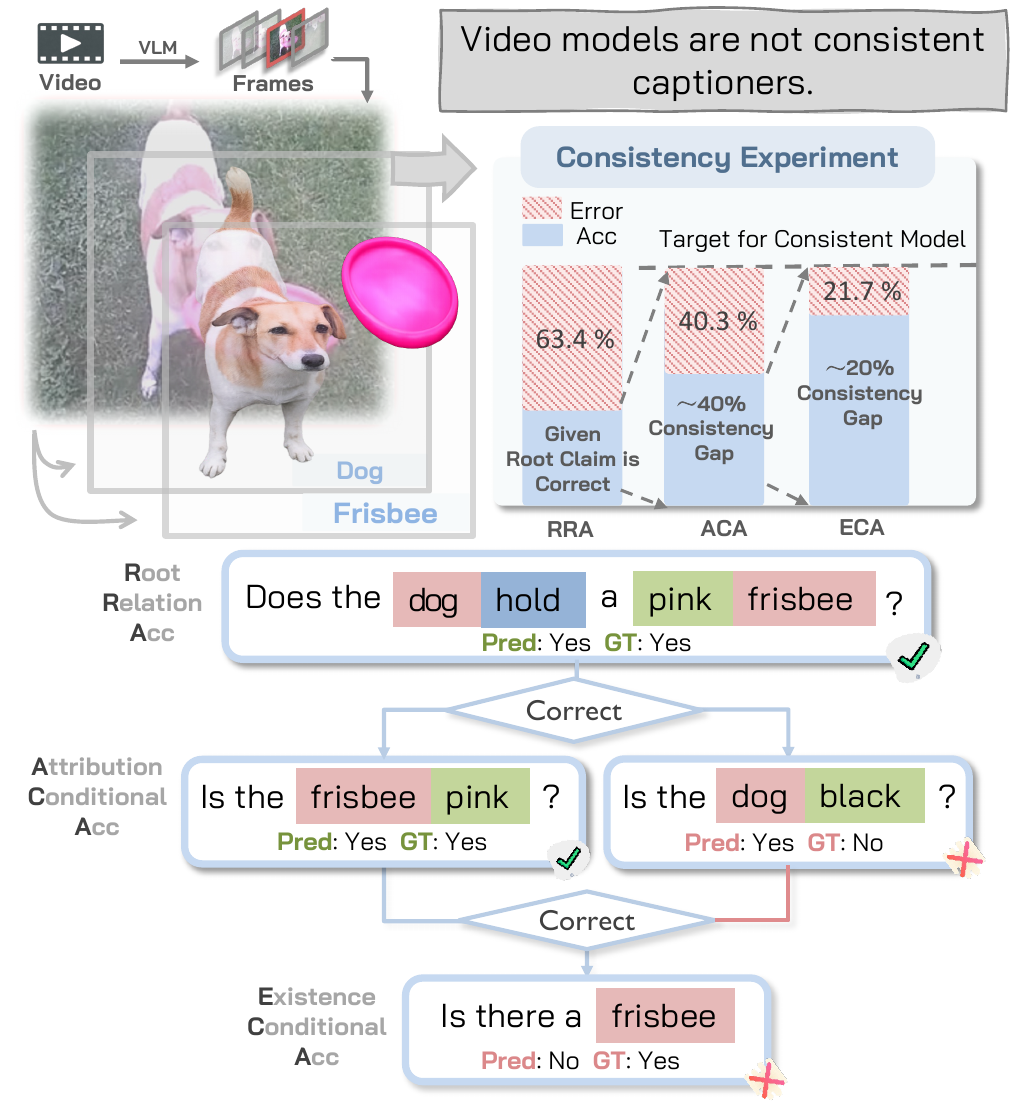}
    \caption{Motivating observation. We evaluate a caption with a top-down compositional consistency audit. Starting from a root claim, we ask whether the supporting evidence beneath it is also correct. The audit set contains 200 human-created audit samples derived from AGQA-Decomp, with the root, attribute, and existence labels manually refined and verified.}
    \label{fig:intro_obs}
    \vspace{-0.8em}
\end{wrapfigure}

\textbf{Top-down audit of caption consistency.} To systematically investigate this failure mode, we shift the evaluation focus from surface sentence plausibility to grounded internal consistency. We conduct a top-down compositional consistency audit, decomposing a caption into a hierarchy of supporting factual and temporal claims. Starting from a root relational claim, we then check whether the prerequisite lower-level evidence is also correct. Let $\mathcal{S}$ denote the set of audit samples, let $\mathcal{S}_{r}=\{i\in\mathcal{S}:\mathrm{Root}_i=\mathrm{GT}_i\}$, and let $\mathcal{S}_{ra}=\{i\in\mathcal{S}:\mathrm{Root}_i=\mathrm{GT}_i \land \mathrm{Attr}_i=\mathrm{GT}_i\}$. Further let $c_i^{r}=\mathbf{1}[\mathrm{Root}_i=\mathrm{GT}_i]$, $c_i^{a}=\mathbf{1}[\mathrm{Attr}_i=\mathrm{GT}_i]$, and $c_i^{e}=\mathbf{1}[\mathrm{Exist}_i=\mathrm{GT}_i]$. We report three empirical accuracies: Root Relation Accuracy (RRA), Attribute Conditional Accuracy (ACA), and Existence Conditional Accuracy (ECA):
\begin{equation*}
\mathrm{RRA}=\mathbb{E}_{i\in\mathcal{S}}[c_i^{r}],\quad
\mathrm{ACA}=\mathbb{E}_{i\in\mathcal{S}_{r}}[c_i^{a}],\quad
\mathrm{ECA}=\mathbb{E}_{i\in\mathcal{S}_{ra}}[c_i^{e}].
\end{equation*}
These three metrics progressively test whether a caption remains correct as we descend the evidence hierarchy. RRA evaluates the top-level relational claim over the full sample set. ACA then asks, among samples with a correct root relation, whether the supporting attribute is also correct. ECA further asks, among samples with both a correct relation and a correct attribute, whether the referenced object existence is correct. Together, this cascade tests whether an apparently correct high-level statement is actually supported by the lower-level evidence beneath it, as illustrated in Figure~\ref{fig:intro_obs}.

\textbf{Video models are not consistent captioners.} Our top-down audit reveals a striking vulnerability in state-of-the-art models: a severe deficit in fine-grained grounding even when high-level predictions are correct. Logically, if a model genuinely comprehends a scene well enough to correctly predict a complex root relational claim, its accuracy on the prerequisite lower-level evidence should approach 100\%. However, our conditional analysis on Qwen3-VL-8B exposes a substantial consistency gap. Even under the favorable condition where the root claim is already correct, the supporting attribute accuracy  only reaches approximately 60\%, and the object-existence accuracy still falls short at around 80\%. Thus, a correct high-level compositional statement does not guarantee that the underlying fine-grained evidence is actively grounded. The model frequently bypasses foundational factual evidence to guess the global context, rendering the reasoning process right for the wrong reasons.

\textbf{Structured consistency via reinforcement learning.} Addressing this deficit requires training signals that better capture whether generated captions are grounded in the video at the level of factual and temporal consistency. We therefore revisit reinforcement learning (RL) through structured reward design. Specifically, we parse generated and reference captions into structured factual and temporal units, and use them to build a sequence-level training objective with three complementary reward branches: (1) a scene-graph reward over objects, attributes, and relations; (2) an instance-aware temporal reward for repeated-event alignment and order consistency; and (3) a video-grounded VQA reward for hierarchical verification. This design better aligns the reward with grounded consistency, while retaining standard sequence-level policy optimization.

Extensive experiments demonstrate that this objective improves caption faithfulness while remaining highly competitive on standard video understanding benchmarks. Across Qwen3-VL backbones, the strongest and most consistent gains are observed on hallucination-sensitive evaluations. This pattern  matches the intended role of our method: structured consistency supervision acts as a powerful mechanism for reducing unsupported content, with downstream performance gains naturally following from this enhanced faithfulness.

In summary, our main contributions are as follows:
\begin{itemize}
    \item We adapt compositional consistency auditing to video captioning, revealing a critical flaw in current VideoLLMs: top-down analysis shows they frequently generate plausible high-level claims that lack valid attribute and existence grounding, functioning as inconsistent visual captioners.
    \item Motivated by this gap, we design structured rewards for sequence-level RL in video captioning. Our objective integrates (1) a instance-aware scene-graph reward, (2) an temporal reward, and (3) a video-grounded VQA reward to explicitly enforce factual and temporal consistency.
    \item We demonstrate that optimizing structured consistency yields robust improvements across multiple open-source backbones. Notably, our approach delivers the most significant gains on hallucination-sensitive evaluations, validating it as a highly practical route to faithful video understanding.
\end{itemize}

\section{Related Work}
\paragraph{Compositional reasoning in video understanding.}
A central idea behind our method is that video understanding should be judged compositionally, not only by a final sentence or answer. This line is most explicit in VideoQA. AGQA introduced compositional spatio-temporal reasoning as a benchmark target, and AGQA-Decomp showed that even when a final answer is correct, its supporting sub-claims may still be inconsistent~\citep{agqa,agqadecomp}. More recent frameworks attempt to expose intermediate structure directly: Video-of-Thought organizes reasoning from perception to cognition, while Align-and-Aggregate parses a question into a compositional graph and resolves the final answer through intermediate sub-questions~\citep{fei2024vot,liao2024alignaggregate}. In parallel, grounded and causal VideoQA methods try to isolate the evidence relevant to each question by alignment or causal refinement~\citep{invariantgrounding,vcsr,cra2025}. Our work is closest in spirit to this literature, but the setting is different: instead of decomposing a \emph{question} at inference time, we decompose a generated \emph{caption} during training and use its scene-graph atoms and ordered events as reward-bearing supervision targets.

\section{Method}
\label{sec:method}

\begin{figure*}[h]
    \centering
    \vspace{-1mm}
    \includegraphics[width=0.975\textwidth]{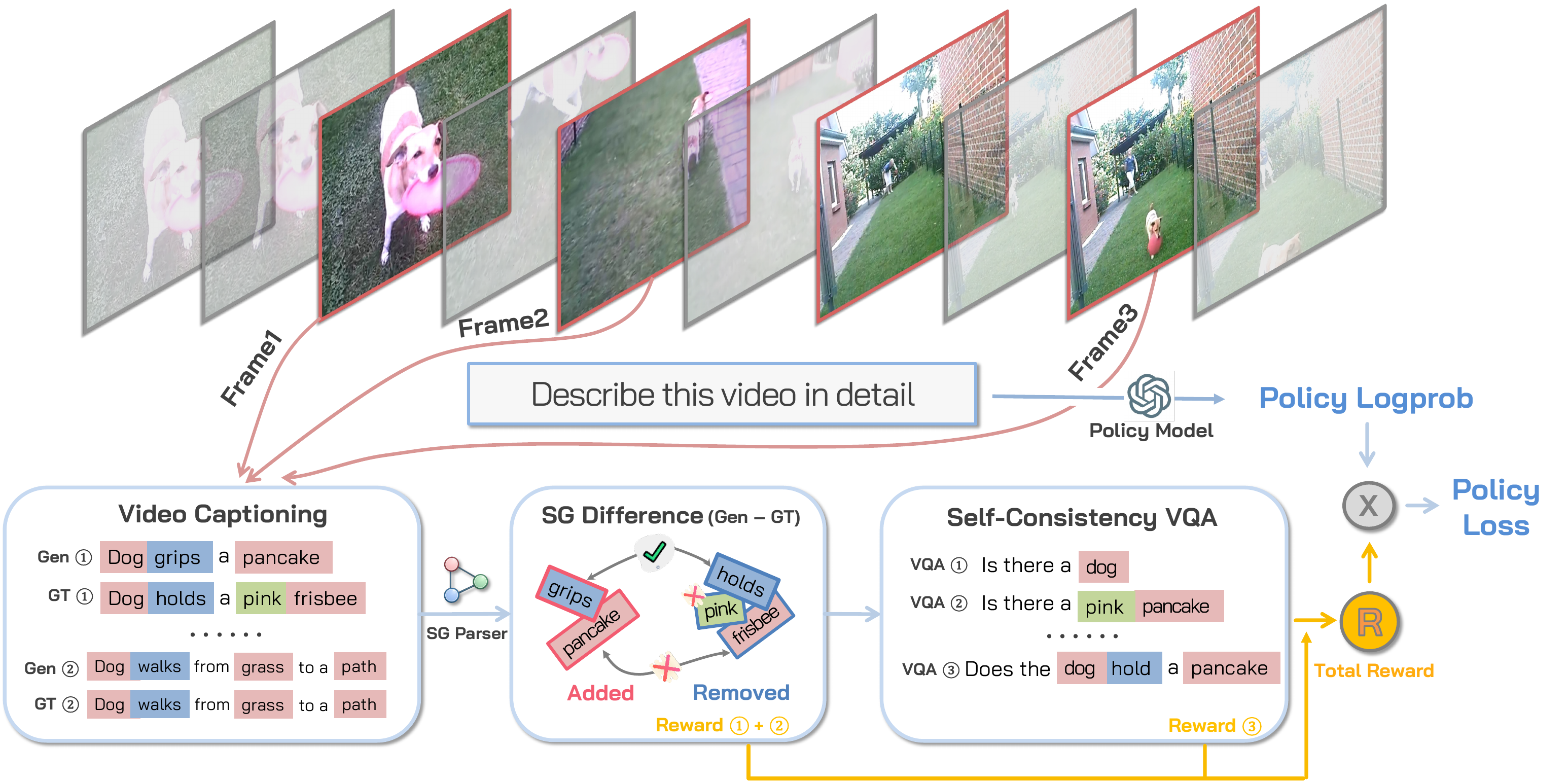}
    \caption{Overview of our method. We compare the sampled caption and the reference caption in a shared structured space rather than at the whole-sentence level. After lightweight preprocessing, both captions are parsed into scene-graph elements, scored by revision-based factual, temporal, and VQA verification rewards, and then used to update the policy with REINFORCE plus KL regularization.}
    \label{fig:mainfig}
\end{figure*}

\subsection{From captions to anchored structured units}
\label{subsec:anchored-units}

Given a video--instruction pair $[V,x]$, the policy samples a caption
$y \sim \pi_\theta(\cdot \mid [V,x])$, where $y^\ast$ denotes the reference caption.
Before computing rewards, we first segment each caption into sentences and clauses using spaCy,
and normalize noun and verb heads by lemmatization~\citep{honnibal2020spacy}. We then apply
the Factual scene-graph parser~\citep{li2023factual} to map both captions into a shared structured
space. Formally, we define a parsing and anchoring function $f_{\mathrm{parse}}$ that extracts a
structured unit set
$U(c) = \{U_{\mathrm{obj}}(c), U_{\mathrm{attr}}(c), U_{\mathrm{rel}}(c), U_{\mathrm{event}}(c)\}$
for any caption $c \in \{y, y^\ast\}$. Since the parser itself is not instance-aware, we further
introduce a deterministic anchoring layer on top of its outputs. Concretely, repeated mentions are
assigned local instance ids, such as \textit{cup$_1$} and \textit{cup$_2$}, and each attribute,
relation, and event mention inherits the ids of the object instances on which it depends. After this
step, each caption is represented as anchored objects, anchored attributes, anchored relations, and
anchored event mentions, together with their participant anchors and clause positions.

These anchors provide the basis for instance awareness: two structured units $u \in U(y)$ and
$u^\ast \in U(y^\ast)$ are considered matchable only when both their semantics and their anchors
are compatible. This design allows the reward to distinguish repeated entities and repeated events
without introducing any additional learned module. Full implementation details are provided in the
Appendix.

\subsection{Structured consistency rewards}
\label{subsec:rewards}

Our reward keeps the same three-branch design, but the three branches now play cleaner and more
distinct roles. The scene-graph branch scores structured revisions between the sampled caption and
the reference caption. The temporal branch and the VQA branch both use the same video-conditioned
verification interface, defined as
$\hat{a}(q) = \arg\max_{a \in \{\texttt{yes}, \texttt{no}\}} \pi(\cdot \mid [V,q])$,
but they supervise different root claims: the temporal branch targets event order and repetition,
while the VQA branch targets factual support chains over objects, attributes, relations, and event
participants.

\paragraph{Revision-based scene-graph reward.}
We first align generated and reference object instances by finding a maximum-weight one-to-one
bipartite matching $\mathcal{M}_{\mathrm{obj}}$ between $U_{\mathrm{obj}}(y)$ and
$U_{\mathrm{obj}}(y^\ast)$, where the edge weight is the semantic similarity between canonicalized
object phrases. This object map $\mathcal{M}_{\mathrm{obj}}$ defines which higher-order units are
allowed to match. An attribute can match only when it is attached to the same aligned object, a
relation can match only when both endpoints are aligned, and an event mention can match only when
its participant anchors are compatible under the same object map. For each unit type
$\tau \in \{\mathrm{obj},\mathrm{attr},\mathrm{rel}\}$, we remove exact overlaps and then solve a
second partial one-to-one matching on the residual units. Let $m_\tau$ denote the resulting matched
mass, including exact overlaps and the similarity of valid residual substitutions. We score each
unit type by
\[
q_\tau = \frac{2m_\tau}{|U_\tau(y)| + |U_\tau(y^\ast)|},
\qquad
q_{\mathrm{sg}} = \sum_{\tau \in \{\mathrm{obj},\mathrm{attr},\mathrm{rel}\}} \alpha_\tau q_\tau,
\]
where $\sum_\tau \alpha_\tau = 1$. If a unit type is absent on both sides, we set its score to $1$.
This revision-based design blocks the main reward-hacking case: once one reference unit has been
claimed by one generated unit, extra near-synonymous additions can no longer collect reward and
only enlarge the denominator.

\paragraph{Temporal reward as temporal root-claim verification.}
We keep temporal as an explicit reward branch, but express it with the same yes/no verification form
as VQA. The key difference from the factual VQA branch is that the root claim is now temporal, not
relational. We first match generated and reference event mentions one-to-one under predicate
similarity after checking participant-anchor compatibility under $\mathcal{M}_{\mathrm{obj}}$. We
formulate a temporal question set $Q_{\mathrm{temp}} = Q_{\mathrm{temp}}^{+} \cup Q_{\mathrm{temp}}^{-}$.
Temporal questions are instantiated only when the reference caption explicitly expresses temporal
structure. Positive temporal questions $Q_{\mathrm{temp}}^{+}$ come from reference-supported event
occurrences and ordered event pairs. Negative temporal questions $Q_{\mathrm{temp}}^{-}$ are
restricted to two clean counterfactuals. The first is \emph{order inversion}: if the reference states
that one matched event happens before another, we also ask the reversed order question. The second
is \emph{instance collapse or swap}: if the sampled caption merges repeated events into the same
actor, or assigns an event to the wrong actor or object anchor, we convert that conflicting temporal
claim into a negative question, but only within the same matched event slots. This keeps the branch
focused on genuine temporal errors rather than open-world omissions.

For each temporal question $q \in Q_{\mathrm{temp}}$, we query the same video-conditioned policy
with $[V,q]$, read a detached one-step answer $\hat{a}(q) \in \{\texttt{yes},\texttt{no}\}$, and
compare it with the known label $a^\ast(q)$. The temporal score is the average binary accuracy:
\[
q_{\mathrm{temp}} = \frac{1}{|Q_{\mathrm{temp}}|} \sum_{q \in Q_{\mathrm{temp}}}
\mathbf{1}[\hat a(q)=a^\ast(q)].
\]

\paragraph{Hierarchical factual VQA reward.}
The VQA branch uses the same binary verification interface, but its goal is different: it checks
whether high-level factual claims are supported by the lower-level evidence underneath them. Starting
from the reference parse, we expand each supported root claim into a short support chain over object
existence, anchor-specific attributes, anchored relations, and event participants. These form the
positive question set $Q_{\mathrm{vqa}}^{+}$. The negative question set $Q_{\mathrm{vqa}}^{-}$ does
not consist of arbitrary facts absent from the reference caption. Instead, we only use
anchor-compatible conflicts at the same structural slot, such as a wrong attribute on a matched
object, a wrong relation between matched objects, or a wrong participant binding exposed by the
sampled caption. This restriction is important because it avoids penalizing valid but unmentioned
video content.

Letting $Q_{\mathrm{vqa}} = Q_{\mathrm{vqa}}^{+} \cup Q_{\mathrm{vqa}}^{-}$, we use the same
detached yes/no answering step as above and define:
\[
q_{\mathrm{vqa}} = \frac{1}{|Q_{\mathrm{vqa}}|} \sum_{q \in Q_{\mathrm{vqa}}}
\mathbf{1}[\hat a(q)=a^\ast(q)].
\]
Detailed templates, filtering rules, and positive/negative construction rules will be shown in the
appendix.

We convert all normalized branch scores into centered rewards with a shared affine form
\[
r_b = \rho\,(q_b-\kappa), \qquad b \in \{\mathrm{sg},\mathrm{temp},\mathrm{vqa}\},
\]
and combine them as
\[
R = \lambda_{\mathrm{sg}} r_{\mathrm{sg}} + \lambda_{\mathrm{temp}} r_{\mathrm{temp}}
+ \lambda_{\mathrm{vqa}} r_{\mathrm{vqa}}.
\]
Here $\kappa$ has the shared centering constant value 0.5 and $\rho$ controls the reward range.
We keep these values fixed in the main experiments and report the search ranges and robustness
ablations in the Appendix.

\subsection{Policy optimization}

We optimize the policy with a REINFORCE-style objective over sampled captions. Let
$\bar{\ell}_{\theta}(y) = \frac{1}{L}\sum_{t=1}^{L}\log \pi_{\theta}(a_t \mid a_{<t}, [V,x])$
be the average log-probability of the sampled caption and let
$\widehat{\mathrm{KL}}(\pi_{\theta}\,\|\,\pi_{\mathrm{ref}})$ be the sampled-token KL against a
frozen reference policy. The training loss is defined over the expectation of the generated samples
$y \sim \pi_\theta$:
\[
\mathcal{L}(\theta)
=
-\mathbb{E}_{y \sim \pi_\theta}\big[R \cdot \bar{\ell}_{\theta}(y)\big]
+ \beta\, \mathbb{E}_{y \sim \pi_\theta}
\big[\widehat{\mathrm{KL}}(\pi_{\theta}\,\|\,\pi_{\mathrm{ref}})\big].
\]

Fundamentally, our design functions as structured reward shaping for sequence-level RL. By
anchoring rewards to key compositional elements, it provides finer-grained supervisory signals than
standard sentence-level RL, while remaining significantly more computationally efficient and stable
than dense token-level reward assignment. All parser, preprocessing, matching, and reward
hyperparameters are fixed across the main experiments and detailed in
Appendix~\ref{app:detail_settings} and Appendix~\ref{app:parser_preproc_settings}.

\section{Experiments}
\subsection{Experimental setup}
\paragraph{Implementation details.}
Unless otherwise specified, we use 8 sampled frames at inference time and keep the evaluation protocol identical across methods. For open-source backbones, we initialize from the corresponding instruction-tuned checkpoint and perform parameter-efficient post-training with LoRA adapters~\citep{hu2022lora}. In the default 8B recipe, we first run SFT on the 1{,}891-sample TemporalBench~\citep{cai2024temporalbench} short-caption set and then apply our RL stage on the aligned training set used by our method, which provides scene-graph, temporal, and QA-style supervision derived from the reference caption. Both stages are optimized with AdamW~\citep{loshchilov2019decoupled}. Appendix~\ref{app:detail_settings} summarizes the default recipe, Appendix~\ref{app:parser_preproc_settings} lists the parser and preprocessing settings used by the reward modules, Appendix~\ref{app:question-construction} summarizes temporal and factual question construction, Appendix~\ref{app:parser_backend_audit} reports the parser-backend audit that motivates our choice of FACTUAL, and Appendix~\ref{app:data_overlap} audits train--evaluation overlap for the TemporalBench-based training pools.
\paragraph{Baselines.}
We compare our method against two categories of baselines:
(i) \textbf{Closed-source VLMs}: GPT-4o~\citep{gpt4} and Gemini 1.5 Pro~\citep{gemini}. 
(ii) \textbf{Open-source VLMs}: LLaVA-OV-7B~\citep{llavavideo}, Qwen2.5-VL-7B~\citep{qwen25vl}, Qwen3-VL-4B~\citep{bai2025qwen3}, and Qwen3-VL-8B. 
For open-source models, we additionally compare against recent RL-based or self-correction-based variants, including Season~\citep{wu2025season}, TPO~\citep{tpo}, RRPO~\citep{rrpo}, DINO-HEAL~\citep{vidhalluc}, and ArrowRL~\citep{arrowrl}. Appendix~\ref{app:other_baseline_details} summarizes the concrete settings used for the reproduced baselines in our comparisons.

\paragraph{Benchmarks.}
We evaluate on three benchmark groups. First, we use \textbf{temporal understanding} benchmarks, TempCompass~\citep{tempcompass} and TVBench~\citep{tvbench}, to test event order and temporal reasoning. Second, we use \textbf{general video understanding} benchmarks, Video-MME~\citep{videomme} and MVBench~\citep{mvbench}, to measure whether gains in faithfulness preserve broad video QA ability. Third, we use \textbf{hallucination-oriented} benchmarks, VideoHallucer~\citep{videohallucer}, and EventHallusion~\citep{eventhallusion}, to probe factual and temporal failure modes more directly. 

\subsection{Main results}
\begin{table}[h]
\centering
\small
\setlength{\tabcolsep}{5pt}
\begin{tabular}{lccc}
\toprule
Metric & Base & Ours & $\Delta$ \\
\midrule
Root Relation Accuracy & 36.6 & 47.8 & +11.2 \\
Attribute Conditional Accuracy & 59.7 & 78.4 & +18.7 \\
Existence Conditional Accuracy & 78.3 & 92.1 & +13.8 \\
\bottomrule
\end{tabular}
\caption{\textbf{Post-training top-down audit under the same protocol as Figure~\ref{fig:intro_obs}.} Our method improves the full root-to-attribute-to-existence support chain, with especially large gains on the conditional attribute and existence accuracies.}
\label{tab:topdown_audit_placeholder}
\end{table}

\paragraph{Top-down audit as a matched diagnostic.}
The top-down audit in Section~1 is also the diagnostic that our method is designed to improve. Table~\ref{tab:topdown_audit_placeholder} applies the same root-to-attribute-to-existence protocol to compare the base model against our post-trained model. The improvement is substantial at every level of the hierarchy. The root relation accuracy increases markedly, which shows that the model becomes more reliable even at the top-level compositional claim. More importantly, the conditional attribute and existence accuracies improve even more, indicating that the gain is not just better global guessing: once the high-level relation is correct, the lower-level support chain is also much more likely to remain grounded. This matched audit closes the loop between motivation and evaluation: the same hierarchy that reveals the original inconsistency gap is also where our method delivers one of its clearest gains.

\begin{table*}[t]
\centering
\small
\setlength{\tabcolsep}{2.6pt}
\renewcommand{\arraystretch}{0.95}
\resizebox{\textwidth}{!}{%
\begin{tabular}{lccccccccc}
\toprule
\multirow{2}{*}{\textbf{Models}} 
& \multicolumn{3}{c}{\textbf{Temporal}} 
& \multicolumn{3}{c}{\textbf{Conventional}} 
& \multicolumn{3}{c}{\textbf{Hallucination}} \\
\cmidrule(lr){2-4} \cmidrule(lr){5-7} \cmidrule(lr){8-10}
& \textbf{TempComp.} & \textbf{TVBench} & \textbf{Avg.}
& \textbf{Video-MME} & \textbf{MVBench} & \textbf{Avg.}
& \textbf{VidHallucer} & \textbf{EventHall.} & \textbf{Avg.} \\
\midrule
\multicolumn{10}{c}{\textbf{\emph{Closed-source Models}}} \\ \midrule
GPT-4o & 73.8 & 39.9 & 56.9 & 71.9 & 49.1 & 60.5 & 53.3 & 91.9 & 72.6 \\
Gemini 1.5 Pro & 67.1 & 47.6 & 57.4 & 75.0 & 60.5 & 67.8 & 37.8 & 80.4 & 59.1 \\
\midrule
\multicolumn{10}{c}{\textbf{\emph{Open-source Models}}} \\ \midrule
LLaVA-OV-7B & 68.3 & 42.8 & 55.6 & 50.6 & 54.4 & 52.5 & 46.4 & 60.2 & 53.3 \\
+TCD & 68.5 & 42.9 & 55.7 & 50.4 & 54.3 & 52.4 & 48.1 & 68.5 & 58.3 \\
+DINO-HEAL & 68.3 & 42.9 & 55.6 & 50.6 & 54.4 & 52.5 & 46.9 & 60.2 & 53.6 \\
+Season & \textbf{69.3} & 43.0 & \cellcolor{babyblue!30}\textbf{56.2} & 50.7 & \textbf{54.8} & 52.8 & \textbf{49.3} & \textbf{69.2} & \cellcolor{babyblue!30}\textbf{59.3} \\
\textbf{+\methodname} & 68.8 & \textbf{43.0} & 55.9 & \textbf{51.2} & 54.6 & \cellcolor{babyblue!30}\textbf{52.9} & 48.5 & 64.2 & 56.4 \\
\midrule
Qwen2.5-VL-7B & 72.9 & 46.7 & 59.8 & 49.9 & 61.8 & 55.9 & 53.1 & 63.3 & 58.2 \\
+TCD & 73.4 & 46.1 & 59.8 & 49.4 & 60.1 & 54.8 & 52.8 & 64.8 & 58.8 \\
+DINO-HEAL & 73.2 & 46.9 & 60.1 & 49.9 & 61.9 & 55.9 & 53.3 & 63.6 & 58.5 \\
+ArrowRL & 72.6 & \textbf{49.4} & 61.0 & 49.6 & 59.8 & 54.7 & 54.5 & \textbf{69.0} & \textbf{61.8} \\
+Season & 73.7 & 47.7 & 60.7 & 50.4 & 62.3 & 56.4 & \textbf{54.8} & 66.0 & 60.4 \\
\textbf{+\methodname} & \textbf{73.8} & 48.6 & \cellcolor{babyblue!30}\textbf{61.2} & \textbf{50.9} & \textbf{62.4} & \cellcolor{babyblue!30}\textbf{56.7} & \textbf{54.8} & 65.8 & 60.3 \\
\midrule
Qwen3-VL-4B & 71.0 & 41.7 & 56.6 & 53.6 & 62.7 & 58.2 & 51.4 & 68.4 & 59.9 \\
+Season & 70.6 & 43.2 & 56.9 & 53.8 & \textbf{64.3} & 59.1 & 54.2 & 67.9 & 61.1 \\
\textbf{+\methodname} & \textbf{73.0} & \textbf{47.1} & \cellcolor{babyblue!30}\textbf{60.1} & \textbf{54.4} & 64.2 & \cellcolor{babyblue!30}\textbf{59.3} & \textbf{55.8} & \textbf{72.0} & \cellcolor{babyblue!30}\textbf{63.9} \\
\midrule
Qwen3-VL-8B & 74.1 & 47.0 & 60.6 & 56.7 & 65.8 & 61.3 & 55.7 & 66.8 & 61.3 \\
+Season & 71.2 & 46.8 & 59.0 & 56.6 & 65.7 & 61.2 & 53.3 & 71.6 & 62.5 \\
\textbf{+\methodname} & \textbf{75.1} & \textbf{48.0} & \cellcolor{babyblue!30}\textbf{61.3} & \textbf{57.4} & \textbf{66.7} & \cellcolor{babyblue!30}\textbf{62.1} & \textbf{56.8} & \textbf{73.4} & \cellcolor{babyblue!30}\textbf{65.1} \\
\midrule
Qwen3-VL-32B & 77.2 & 51.3 & 64.3 & 60.1 & 69.5 & 64.8 & 62.0 & 77.0 & 69.5 \\
+Season & 74.1 & 49.5 & 61.8 & \textbf{60.9} & 69.6 & \cellcolor{babyblue!30}\textbf{65.3} & 55.1 & 71.6 & 63.4 \\
\textbf{+\methodname} & \textbf{77.3} & \textbf{51.4} & \cellcolor{babyblue!30}\textbf{64.4} & 60.3 & \textbf{69.8} & 65.1 & \textbf{62.5} & \textbf{78.0} & \cellcolor{babyblue!30}\textbf{70.3} \\
\bottomrule
\end{tabular}%
}
\vspace{1mm}
\caption{Performance comparisons on benchmarks for temporal, conventional, and hallucination-oriented video understanding. \textbf{Bold} marks the best per group; highlights indicate the best benchmark results.}
\label{tab:main_video_all}
\end{table*}

\paragraph{General video understanding evaluation.}Table~\ref{tab:main_video_all} reports results on temporal and conventional video-understanding benchmarks. The main pattern is that our training objective generalizes across multiple open-source backbones, rather than benefiting only a single model family. On LLaVA-OV-7B, our method yields a modest but consistent improvement over the base model and remains competitive with recent state of art methods. On Qwen2.5-VL-7B, it achieves the best average performance within the group. The same trend is even clearer on Qwen3-VL backbones, it improves on all four benchmarks simultaneously. Overall, these results suggest that the proposed rewards improve temporal faithfulness without sacrificing broader video-understanding ability.

\paragraph{Hallucination benchmark evaluation.}Table~\ref{tab:main_video_all} also shows that the gains are more pronounced on hallucination-oriented evaluation. For Qwen2.5-VL-7B, our method improves the reported VideoHallucer score over prior baselines and remains competitive on EventHallusion. For both Qwen3-VL-4B and Qwen3-VL-8B, it substantially improves VideoHallucer, and EventHallusion over the corresponding base models. This pattern indicates that the structured reward strengthens factual grounding and temporal consistency.

\subsection{Ablation study}
We ablate the method along four axes: reward components, reward weights, training data scale, and frame sampling strategy. These ablations test complementary questions: which reward branch is necessary, whether the method is sensitive to reweighting, whether the improvement depends on large-scale training data, and whether the gains can be explained simply by changing the visual sampling budget.

\begin{figure*}[h]
\centering
\vspace{-1mm}
\includegraphics[width=0.975\textwidth]{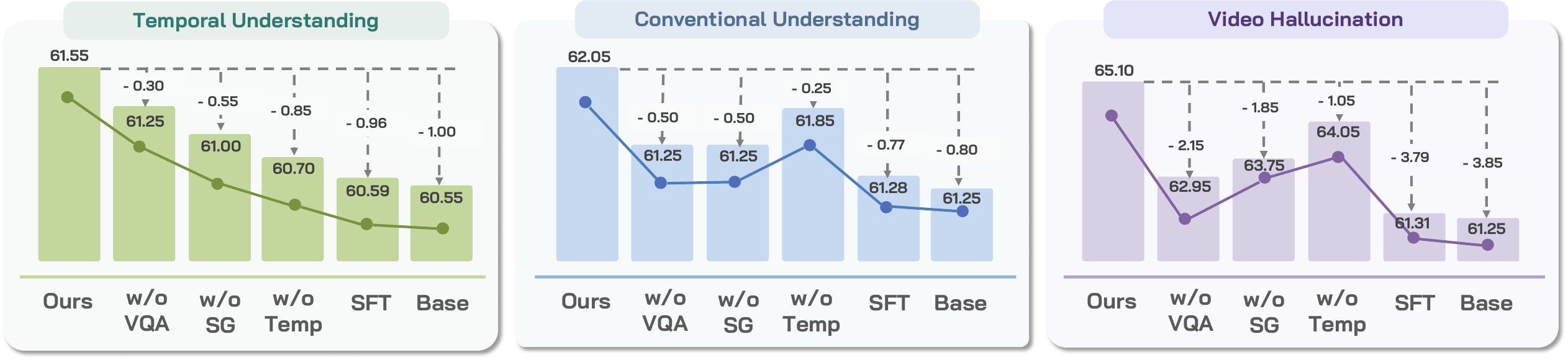}
\caption{\textbf{Reward-component and training-stage ablations.} Results are averaged within temporal understanding, conventional video understanding, and hallucination-oriented benchmark groups.}
\label{fig:component_ablation}
\end{figure*}

\textbf{Ablation on reward components.}
Figure~\ref{fig:component_ablation} summarizes component and training-stage ablations by averaging benchmarks within temporal understanding, conventional video understanding, and hallucination-oriented evaluation. The full model is best in all three groups, which shows that the gain does not come from one dominant branch alone. Removing the scene-graph reward mainly hurts temporal and conventional understanding, while removing the VQA reward causes the largest drop on hallucination-oriented benchmarks. Removing the temporal reward also degrades the result, although the drop is smaller than removing the other two branches. The same figure shows that both the SFT model and the raw backbone remain clearly below the full RL system, so the final gain is not explained by supervised initialization alone. For the corresponding numeric table, see Appendix~\ref{app:additional_ablation_tables}.

\begin{table}[h]
   \centering
   \fontsize{8pt}{9pt}\selectfont
   \resizebox{\textwidth}{!}{%
   \begin{tabular}{lccccccc}
      \toprule
         \textbf{Setting} & $\mathbf{w_{\text{SG}}}$ & $\mathbf{w_{\text{Temp}}}$ & $\mathbf{w_{\text{VQA}}}$ & \textbf{Temporal} & \textbf{Conventional} & \textbf{Hallucination} & \textbf{Avg.} \\
         \midrule
         Ours & 0.15 & 0.25 & 0.35 & 61.55 & 62.05 & 65.10 & 62.90 \\
         Rebalance to SG & 0.20 & 0.25 & 0.35 & 61.43 & 62.10 & 65.22 & 62.92 \\
         Rebalance to Temp & 0.15 & 0.30 & 0.35 & 61.70 & 61.96 & 65.00 & 62.89 \\
         Rebalance to VQA & 0.15 & 0.25 & 0.40 & 61.31 & 62.14 & 65.21 & 62.89 \\
         \bottomrule
      \end{tabular}%
   }
   \caption{\textbf{Ablation study on reward weights for Qwen3-VL-8B.} SG refers to the scene graph reward. Results are averaged within temporal understanding, conventional video understanding, and hallucination-oriented benchmark groups.}
   \label{tab:ablation_components_weight}
\end{table}

\textbf{Ablation on reward weight.}
Table~\ref{tab:ablation_components_weight} studies how sensitive the method is to reward reweighting on Qwen3-VL-8B. The main observation is that the method is robust: across all completed settings, the group-level average stays in a very narrow band so moderate perturbations of the three weights do not materially change the final result. The changes that do appear are also intuitive. Increasing the temporal weight gives the strongest temporal-group score, while increasing the VQA weight slightly favors the hallucination-oriented and conventional groups. Reweighting toward the base reward gives the highest overall average by a negligible margin, which suggests that no single branch needs to dominate the objective. Overall, the gain comes from combining all supervision signals rather than from fragile tuning of one particular weight.

\begin{figure*}[h]
\centering
\vspace{-1mm}
\includegraphics[width=0.995\textwidth]{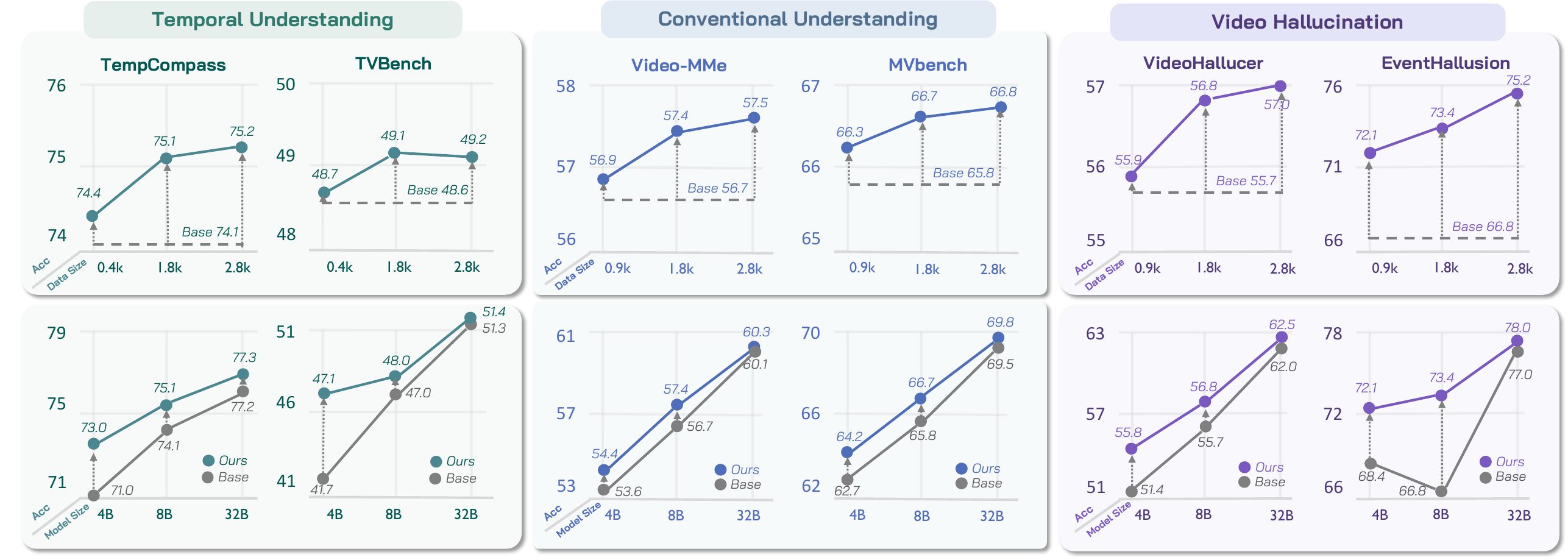}
\caption{\textbf{Scaling behavior across model size and training data size.}
We study two axes of scaling for our post-training recipe: backbone size and training data size. Results are averaged within temporal understanding, conventional video understanding, and hallucination-oriented benchmark groups.}
\label{fig:scaling}
\end{figure*}

\textbf{Ablation on data scaling.}
Figure~\ref{fig:scaling} shows a clear data-scaling trend: as the amount of training data increases, performance also improves across temporal understanding, conventional video QA, and hallucination-oriented evaluation. 
At the same time, the improvement exhibits diminishing returns. Increasing the training set from the smallest to the medium scale brings the most visible gains, while further scaling to the largest setting still improves performance but by a smaller margin. The effect is most pronounced on hallucination-sensitive benchmarks, with smaller yet still positive gains on temporal and conventional benchmarks. This suggests that additional training data continues to strengthen factual grounding and temporal consistency, although the marginal utility gradually decreases as the training set grows. For the complete results, see Appendix~\ref{app:additional_scaling}, Table~\ref{tab:ablation_model_size}.

\textbf{Ablation on frame sampling strategy.}
Table~\ref{tab:ablation_frame_sampling} demonstrates that our method achieves larger performance gains when increasing the frame count from 8 to 32, indicating its ability to leverage additional visual evidence effectively. At 64 frames, however, the marginal improvement diminishes as extra frames become redundant, weakly relevant, or distracting. Overall, while a higher frame budget enhances observability, the benefits saturate when additional context adds little valuable information. Notably, the performance gap between our method and the base model remains wider at higher frame budgets, confirming that our post-training objective enables more reliable utilization of visual evidence.

\begin{table*}[h]
\centering
\footnotesize
\setlength{\tabcolsep}{8pt}
\begin{tabular}{llcccccc}
\toprule
\multirow{2}{*}{\textbf{Category}} & \multirow{2}{*}{\textbf{Benchmark}} & \multicolumn{3}{c}{\textbf{Base}} & \multicolumn{3}{c}{\textbf{Ours}} \\
\cmidrule(lr){3-5}\cmidrule(lr){6-8}
& & \textbf{8} & \textbf{32} & \textbf{64} & \textbf{8} & \textbf{32} & \textbf{64} \\
\midrule
\multirow{2}{*}{\textbf{Temporal}} 
& TempCompass    & 74.1 & 74.6 & 74.2 & 75.1$_{\scriptstyle +1.0}$ & 75.5$_{\scriptstyle +0.9}$ & 76.4$_{\scriptstyle +2.2}$ \\
& TVBench        & 47.0 & 46.8 & 47.9 & 48.0$_{\scriptstyle +1.0}$ & 48.7$_{\scriptstyle +1.9}$ & 49.2$_{\scriptstyle +1.3}$ \\
\midrule
\multirow{2}{*}{\textbf{Conventional}} 
& Video-MME      & 56.7 & 64.2 & 67.0 & 57.4$_{\scriptstyle +0.7}$ & 64.6$_{\scriptstyle +0.4}$ & 67.3$_{\scriptstyle +0.3}$ \\
& MVBench        & 65.8 & 68.5 & 69.2 & 66.7$_{\scriptstyle +0.9}$ & 69.1$_{\scriptstyle +0.6}$ & 69.4$_{\scriptstyle +0.2}$ \\
\midrule
\multirow{2}{*}{\textbf{Hallucination}} 
& VideoHallucer  & 55.7 & 60.3 & 62.2 & 56.6$_{\scriptstyle +0.9}$ & 67.2$_{\scriptstyle +6.9}$ & 63.1$_{\scriptstyle +0.9}$ \\
& EventHallusion & 66.8 & 72.9 & 72.6 & 72.9$_{\scriptstyle +6.1}$ & 75.1$_{\scriptstyle +2.2}$ & 73.1$_{\scriptstyle +0.5}$ \\
\bottomrule
\end{tabular}
\caption{\textbf{Frame-budget ablations on Qwen3-VL-8B.} Increasing sampled frames from 8 to 32 yields substantial gains across most benchmarks, while further scaling to 64 frames leads to performance saturation or slight degradation due to redundancy and noise.}
\label{tab:ablation_frame_sampling}
\end{table*}
\section{Conclusion}
In this work, we investigate an overlooked failure mode in video MLLMs: models frequently generate fluent and contextually plausible captions that lack faithful visual grounding. Our top-down compositional consistency audit reveals that even when high-level relational claims appear correct, they are often unsupported by valid underlying attribute and object existence evidence, highlighting the fundamental limitation of relying solely on standard sentence-level supervision. To address this issue, we propose a sequence-level reinforcement learning framework that enforces structured consistency by incorporating instance-aware scene-graph rewards, temporal  rewards, and hierarchical VQA verification. Extensive evaluations across temporal, general, and hallucination-sensitive benchmarks demonstrate that our approach significantly reduces object-behavior hallucinations and improves factual faithfulness without sacrificing general video reasoning performance. Overall, our findings underscore structured consistency as a key principle for building robust and trustworthy video language models, offering a promising foundation for future post-training strategies.

\section*{Acknowledgements}
This work was supported in part by Advanced Micro Devices, Inc. under the AMD University Program’s AI \& HPC Cluster.

\bibliography{colm2026_conference}

@article{wu2025season,
  title={SEASON: Mitigating Temporal Hallucination in Video Large Language Models via Self-Diagnostic Contrastive Decoding},
  author={Wu, Chang-Hsun and Chang, Kai-Po and Sheng, Yu-Yang and Chung, Hung-Kai and Wang, Kuei-Chun and Wang, Yu-Chiang Frank},
  journal={arXiv preprint arXiv:2512.04643},
  year={2025}
}

@inproceedings{vidhalluc,
  title={Vidhalluc: Evaluating temporal hallucinations in multimodal large language models for video understanding},
  author={Li, Chaoyu and Im, Eun Woo and Fazli, Pooyan},
  booktitle={Proceedings of the Computer Vision and Pattern Recognition Conference},
  pages={13723--13733},
  year={2025}
}

@article{videohallucer,
  title={Videohallucer: Evaluating intrinsic and extrinsic hallucinations in large video-language models},
  author={Wang, Yuxuan and Wang, Yueqian and Zhao, Dongyan and Xie, Cihang and Zheng, Zilong},
  journal={arXiv preprint arXiv:2406.16338},
  year={2024}
}

@article{eventhallusion,
  title={Eventhallusion: Diagnosing event hallucinations in video llms},
  author={Zhang, Jiacheng and Jiao, Yang and Chen, Shaoxiang and Zhao, Na and Tan, Zhiyu and Li, Hao and Ma, Xingjun and Chen, Jingjing},
  journal={arXiv preprint arXiv:2409.16597},
  year={2024}
}

@article{llavavideo,
  title={Llava-video: Video instruction tuning with synthetic data},
  author={Zhang, Yuanhan and Wu, Jinming and Li, Wei and Li, Bo and Ma, Zejun and Liu, Ziwei and Li, Chunyuan},
  journal={arXiv preprint arXiv:2410.02713},
  year={2024}
}

@article{qwen25vl,
  title={Qwen2.5-VL Technical Report},
  author={Bai, Shuai and Chen, Keqin and Liu, Xuejing and Wang, Jialin and Ge, Wenbin and Song, Sibo and Dang, Kai and Wang, Peng and Wang, Shijie and Tang, Jun and others},
  journal={arXiv preprint arXiv:2502.13923},
  year={2025}
}

@inproceedings{tempcompass,
  title={Tempcompass: Do video llms really understand videos?},
  author={Liu, Yuanxin and Li, Shicheng and Liu, Yi and Wang, Yuxiang and Ren, Shuhuai and Li, Lei and Chen, Sishuo and Sun, Xu and Hou, Lu},
  booktitle={Findings of the Association for Computational Linguistics: ACL 2024},
  pages={8731--8772},
  year={2024}
}

@article{tvbench,
  title={Lost in Time: A New Temporal Benchmark for VideoLLMs},
  author={Cores, Daniel and Dorkenwald, Michael and Mucientes, Manuel and Snoek, Cees GM and Asano, Yuki M},
  journal={arXiv preprint arXiv:2410.07752},
  year={2024}
}

@inproceedings{videomme,
  title={Video-mme: The first-ever comprehensive evaluation benchmark of multi-modal llms in video analysis},
  author={Fu, Chaoyou and Dai, Yuhan and Luo, Yongdong and Li, Lei and Ren, Shuhuai and Zhang, Renrui and Wang, Zihan and Zhou, Chenyu and Shen, Yunhang and Zhang, Mengdan and others},
  booktitle={Proceedings of the Computer Vision and Pattern Recognition Conference},
  pages={24108--24118},
  year={2025}
}

@inproceedings{mvbench,
  title={Mvbench: A comprehensive multi-modal video understanding benchmark},
  author={Li, Kunchang and Wang, Yali and He, Yinan and Li, Yizhuo and Wang, Yi and Liu, Yi and Wang, Zun and Xu, Jilan and Chen, Guo and Luo, Ping and others},
  booktitle={Proceedings of the IEEE/CVF Conference on Computer Vision and Pattern Recognition},
  pages={22195--22206},
  year={2024}
}

@article{bai2025qwen3,
  title={Qwen3-vl technical report},
  author={Bai, Shuai and Cai, Yuxuan and Chen, Ruizhe and Chen, Keqin and Chen, Xionghui and Cheng, Zesen and Deng, Lianghao and Ding, Wei and Gao, Chang and Ge, Chunjiang and others},
  journal={arXiv preprint arXiv:2511.21631},
  year={2025}
}

@article{gpt4,
  title={Gpt-4 technical report},
  author={Achiam, Josh and Adler, Steven and Agarwal, Sandhini and Ahmad, Lama and Akkaya, Ilge and Aleman, Florencia Leoni and Almeida, Diogo and Altenschmidt, Janko and Altman, Sam and Anadkat, Shyamal and others},
  journal={arXiv preprint arXiv:2303.08774},
  year={2023}
}

@article{gemini,
  title={Gemini: a family of highly capable multimodal models},
  author={Team, Gemini and Anil, Rohan and Borgeaud, Sebastian and Alayrac, Jean-Baptiste and Yu, Jiahui and Soricut, Radu and Schalkwyk, Johan and Dai, Andrew M and Hauth, Anja and Millican, Katie and others},
  journal={arXiv preprint arXiv:2312.11805},
  year={2023}
}

@article{tpo,
  title={Temporal preference optimization for long-form video understanding},
  author={Li, Rui and Wang, Xiaohan and Zhang, Yuhui and Zohar, Orr and Wang, Zeyu and Yeung-Levy, Serena},
  journal={arXiv preprint arXiv:2501.13919},
  year={2025}
}

@article{rrpo,
  title={Self-alignment of large video language models with refined regularized preference optimization},
  author={Sarkar, Pritam and Etemad, Ali},
  journal={arXiv preprint arXiv:2504.12083},
  year={2025}
}

@article{arrowrl,
  title={Seeing the Arrow of Time in Large Multimodal Models},
  author={Xue, Zihui and Luo, Mi and Grauman, Kristen},
  journal={arXiv preprint arXiv:2506.03340},
  year={2025}
}

@inproceedings{agqa,
  title={AGQA: A Benchmark for Compositional Spatio-Temporal Reasoning},
  author={Grunde-McLaughlin, Madeleine and Krishna, Ranjay and Agrawala, Maneesh},
  booktitle={Proceedings of the IEEE/CVF Conference on Computer Vision and Pattern Recognition},
  pages={11287--11297},
  year={2021}
}

@inproceedings{agqadecomp,
  title={Measuring compositional consistency for video question answering},
  author={Gandhi, Mona and Gul, Mustafa Omer and Prakash, Eva and Grunde-McLaughlin, Madeleine and Krishna, Ranjay and Agrawala, Maneesh},
  booktitle={Proceedings of the IEEE/CVF Conference on Computer Vision and Pattern Recognition},
  pages={5046--5055},
  year={2022}
}

@inproceedings{fei2024vot,
  title={Video-of-Thought: Step-by-Step Video Reasoning from Perception to Cognition},
  author={Fei, Hao and Wu, Shengqiong and Ji, Wenbo and Zhang, Hua and Zhang, Meishan and Lee, Mong Li and Hsu, Winston},
  booktitle={International Conference on Machine Learning},
  year={2024}
}

@inproceedings{liao2024alignaggregate,
  title={Align and Aggregate: Compositional Reasoning with Video Alignment and Answer Aggregation for Video Question-Answering},
  author={Liao, Zhaohe and Li, Jiangtong and Niu, Li and Zhang, Liqing},
  booktitle={Proceedings of the IEEE/CVF Conference on Computer Vision and Pattern Recognition},
  pages={13395--13404},
  year={2024}
}

@inproceedings{invariantgrounding,
  title={Invariant grounding for video question answering},
  author={Li, Yicong and Wang, Xiang and Xiao, Junbin and Ji, Wei and Chua, Tat-Seng},
  booktitle={Proceedings of the IEEE/CVF Conference on Computer Vision and Pattern Recognition},
  pages={2928--2937},
  year={2022}
}

@inproceedings{vcsr,
  title={Visual causal scene refinement for video question answering},
  author={Wei, Yushen and Liu, Yang and Yan, Hong and Li, Guanbin and Lin, Liang},
  booktitle={Proceedings of the 31st ACM International Conference on Multimedia},
  pages={377--386},
  year={2023}
}

@inproceedings{cra2025,
  title={Cross-modal causal relation alignment for video question grounding},
  author={Chen, Weixing and Liu, Yang and Chen, Binglin and Su, Jiandong and Zheng, Yongsen and Lin, Liang},
  booktitle={Proceedings of the Computer Vision and Pattern Recognition Conference},
  pages={24087--24096},
  year={2025}
}

@article{llava,
  title={Visual instruction tuning},
  author={Liu, Haotian and Li, Chunyuan and Wu, Qingyang and Lee, Yong Jae},
  journal={Advances in neural information processing systems},
  volume={36},
  pages={34892--34916},
  year={2023}
}

@article{qwenvl,
  title={Qwen2-vl: Enhancing vision-language model's perception of the world at any resolution},
  author={Wang, Peng and Bai, Shuai and Tan, Sinan and Wang, Shijie and Fan, Zhihao and Bai, Jinze and Chen, Keqin and Liu, Xuejing and Wang, Jialin and Ge, Wenbin and others},
  journal={arXiv preprint arXiv:2409.12191},
  year={2024}
}

@inproceedings{imagecap,
  title={Image captioning with semantic attention},
  author={You, Quanzeng and Jin, Hailin and Wang, Zhaowen and Fang, Chen and Luo, Jiebo},
  booktitle={Proceedings of the IEEE conference on computer vision and pattern recognition},
  pages={4651--4659},
  year={2016}
}

@inproceedings{msrvtt,
  title={Msr-vtt: A large video description dataset for bridging video and language},
  author={Xu, Jun and Mei, Tao and Yao, Ting and Rui, Yong},
  booktitle={Proceedings of the IEEE conference on computer vision and pattern recognition},
  pages={5288--5296},
  year={2016}
}

@inproceedings{vqa,
  title={Vqa: Visual question answering},
  author={Antol, Stanislaw and Agrawal, Aishwarya and Lu, Jiasen and Mitchell, Margaret and Batra, Dhruv and Zitnick, C Lawrence and Parikh, Devi},
  booktitle={Proceedings of the IEEE international conference on computer vision},
  pages={2425--2433},
  year={2015}
}

@inproceedings{activitynetqa,
  title={Activitynet-qa: A dataset for understanding complex web videos via question answering},
  author={Yu, Zhou and Xu, Dejing and Yu, Jun and Yu, Ting and Zhao, Zhou and Zhuang, Yueting and Tao, Dacheng},
  booktitle={Proceedings of the AAAI Conference on Artificial Intelligence},
  volume={33},
  number={01},
  pages={9127--9134},
  year={2019}
}

@inproceedings{hu2022lora,
  title={LoRA: Low-rank adaptation of large language models},
  author={Hu, Edward J. and Shen, Yelong and Wallis, Phillip and Allen-Zhu, Zeyuan and Li, Yuanzhi and Wang, Shean and Wang, Lu and Chen, Weizhu},
  booktitle={International Conference on Learning Representations},
  year={2022}
}

@inproceedings{loshchilov2019decoupled,
  title={Decoupled weight decay regularization},
  author={Loshchilov, Ilya and Hutter, Frank},
  booktitle={International Conference on Learning Representations},
  year={2019}
}

@inproceedings{li2023factual,
  title={{FACTUAL}: A Benchmark for Faithful and Consistent Textual Scene Graph Parsing},
  author={Li, Zhuang and Chai, Yuyang and Zhuo, Terry Yue and Qu, Lizhen and Haffari, Gholamreza and Li, Fei and Ji, Donghong and Tran, Quan Hung},
  booktitle={Findings of the Association for Computational Linguistics: ACL 2023},
  pages={6377--6390},
  year={2023},
  url={https://aclanthology.org/2023.findings-acl.398}
}

@inproceedings{reimers2019sentencebert,
  title={Sentence-BERT: Sentence Embeddings using Siamese BERT-Networks},
  author={Reimers, Nils and Gurevych, Iryna},
  booktitle={Proceedings of the 2019 Conference on Empirical Methods in Natural Language Processing},
  pages={3982--3992},
  year={2019},
  url={https://aclanthology.org/D19-1410}
}

@misc{honnibal2020spacy,
  author = {Honnibal, Matthew and Montani, Ines and Van Landeghem, Sofie and Boyd, Adriane},
  title = {spaCy: Industrial-strength Natural Language Processing in Python},
  year = {2020},
  publisher = {Zenodo},
  doi = {10.5281/zenodo.1212303},
  url = {https://doi.org/10.5281/zenodo.1212303}
}

@misc{openai_api_pricing_2026,
  author = {{OpenAI}},
  title = {OpenAI API Pricing},
  year = {2026},
  howpublished = {\url{https://openai.com/api/pricing/}},
  note = {Accessed: 2026-03-31}
}

@article{cai2024temporalbench,
  title={Temporalbench: Benchmarking fine-grained temporal understanding for multimodal video models},
  author={Cai, Mu and Tan, Reuben and Zhang, Jianrui and Zou, Bocheng and Zhang, Kai and Yao, Feng and Zhu, Fangrui and Gu, Jing and Zhong, Yiwu and Shang, Yuzhang and others},
  journal={arXiv preprint arXiv:2410.10818},
  year={2024}
}
\bibliographystyle{colm2026_conference}

\newpage
\appendix
\section*{Appendix}
\setcounter{section}{0}
\setcounter{subsection}{0}
\setcounter{subsubsection}{0}
\renewcommand{\thesection}{}
\renewcommand{\thesubsection}{\arabic{subsection}}
\renewcommand{\thesubsubsection}{\arabic{subsection}.\arabic{subsubsection}}

This appendix is organized into three parts. We first summarize the default recipe and parser configuration used in the main experiments. We then analyze two practical aspects of the training pipeline: why we use FACTUAL as the default parser and why the TemporalBench-based training pool does not create direct benchmark leakage. Finally, we collect additional captioning, scaling, and ablation results that complement the main text.

\subsection{Declaration of LLM use}
In preparing this submission, we used LLMs as writing and editing assistants for language refinement, grammar correction, formatting cleanup, and presentation polishing. They were also used to help inspect bibliography metadata and LaTeX consistency under author supervision. We did not use LLMs to generate experimental results or to replace author judgment in problem formulation, methodology, implementation, or evaluation. All substantive research decisions, experiments, and final validation were carried out and verified by the authors.

\subsection{Training Recipe and Parser Settings}

\subsubsection{Detail Settings}
\label{app:detail_settings}
Our default 8B recipe uses a simple two-stage schedule. We first apply LoRA SFT~\citep{hu2022lora} on the 1{,}891-sample TemporalBench short-caption set and then continue with LoRA RL on the aligned training set used by our method, derived from the same caption pool. This design keeps the underlying video domain fixed across stages and changes only the supervision granularity, so the gains in the main paper are not attributable to a shift in training videos or prompt format.

Table~\ref{tab:detail_settings} summarizes the default configuration used in the main 8B runs. In the data-scale ablation, we enlarge only the RL-stage caption pool while keeping the backbone, SFT initialization, and evaluation protocol fixed; this makes the scaling trend in Section~4 easier to interpret as a supervision-scale effect rather than a training-recipe change.

\begin{table}[h]
\centering
\small
\setlength{\tabcolsep}{4pt}
\resizebox{\columnwidth}{!}{%
\begin{tabular}{lcc}
\toprule
\textbf{Setting} & \textbf{SFT} & \textbf{RL} \\
\midrule
Backbone & Qwen3-VL-8B-Instruct & SFT-initialized Qwen3-VL-8B \\
Training data & TemporalBench short-caption & Aligned SFT version of the same pool \\
\# training samples & 1,891 & 1,891 \\
Fine-tuning & LoRA ($r{=}8$, $\alpha{=}16$, drop.=0.05) & LoRA ($r{=}8$, $\alpha{=}16$, drop.=0.05) \\
Optimizer & AdamW & AdamW \\
Learning rate & $1\times10^{-5}$ & $5\times10^{-6}$ \\
Schedule & cosine, warmup 5\% & cosine, warmup 3\% \\
Batch / accum. & 1 / 2 & 1 / 4 \\
Context length & 8192 & 8192 \\
Epochs / steps & 1 epoch & 300 steps \\
Rollout decoding & -- & one-turn, $T{=}1.0$, $p{=}0.9$, max\_new\_tokens=256 \\
Reward weights & -- & $(w_{\text{sg}}, w_{\text{temp}}, w_{\text{vqa}}) = (0.15, 0.25, 0.35)$ \\
\bottomrule
\end{tabular}%
}
\caption{Default 8B training recipe used in the main results.}
\label{tab:detail_settings}
\end{table}

\subsubsection{Parser and Preprocessing Settings}
\label{app:parser_preproc_settings}
Our reward modules use a fixed parser stack throughout all main experiments. For scene-graph extraction, we use the public FACTUAL parser checkpoint~\citep{li2023factual} `lizhuang144/flan-t5-base-VG-factual-sg' and parse sentences independently before merging sentence-level outputs. This choice gives us a stable object--attribute--relation interface for the factual reward without introducing another trainable component into the RL loop.

For preprocessing, we use spaCy~\citep{honnibal2020spacy} with the English pipeline `en\_core\_web\_sm' to perform sentence splitting and identify noun- and verb-centered phrases and instance indexing. Repeated mentions are indexed so that downstream rewards can distinguish, for example, `cup\_1' from `cup\_2' or two repeated event mentions of the same predicate. For soft semantic comparison, we encode structured atoms with `sentence-transformers/all-mpnet-base-v2' following the Sentence-BERT framework~\citep{reimers2019sentencebert}. Temporal preprocessing then applies lightweight regex-based cleanup before event extraction and bucket construction. Table~\ref{tab:parser_preproc_settings} lists the concrete parser, preprocessing, and decoding parameters used in the main runs.

\begin{table*}[t]
\centering
\footnotesize
\setlength{\tabcolsep}{8pt}
\begin{tabular}{ll}
\toprule
\textbf{Module / Parameter} & \textbf{Setting} \\
\midrule
Scene-graph parser & FACTUAL parser (\texttt{flan-t5-base-VG-factual-sg}) \\
Beam size & 5 \\
Max input length & 512 \\
Max output length & 128 \\
Sentence preprocessing & spaCy (\texttt{en\_core\_web\_sm}) sentence splitting \\
\bottomrule
\end{tabular}
\caption{Parser and preprocessing settings used by the reward modules in the main experiments.}
\label{tab:parser_preproc_settings}
\end{table*}

\subsection{Question construction for temporal and factual verification}
\label{app:question-construction}

The temporal and factual verification branches use the same answer interface and the same policy, but their question sources are different. Temporal questions are built only from reference event occurrences and reference event pairs that explicitly encode order or repetition. Factual questions are built from local support chains under reference objects, attributes, relations, and event participants. In both branches, the answer label is known by construction, so no external verifier is needed.

Positive questions always come from the reference parse. Negative questions are created only when the sampled caption changes a matched slot. For the factual branch, this means a wrong attribute on a matched object, a wrong relation between matched objects, or a wrong participant binding on a matched event slot. For the temporal branch, this means either a reversed supported order or a collapse/swap error where repeated events are assigned to the wrong matched actor or object anchors. We never create negatives from arbitrary extra nouns or arbitrary facts that are merely unmentioned in the reference caption.

Table~\ref{tab:qa-templates} gives the fixed template inventory used in the revised method. In the main setting, we keep the wording fixed across all experiments and score only the first answer token over \texttt{yes}/\texttt{no}. If a hard question budget is needed, we apply an optional cap after balancing positive and negative questions.

\begin{table}[h]
\centering
\small
\resizebox{\columnwidth}{!}{%
\begin{tabular}{llll}
\toprule
Root type & Template & Positive source & Negative source \\
\midrule
Object existence & \texttt{Is there a/an [OBJECT]?} & reference support chain & none; we do not create open-world existence negatives \\
Attribute & \texttt{Is the [OBJECT] [ATTRIBUTE]?} & reference support chain & wrong attribute on a matched object slot \\
Relation & \texttt{Does the [SUBJECT] [RELATION] the [OBJECT]?} & reference support chain & wrong relation on matched endpoints \\
Event occurrence & \texttt{Does [PARTICIPANTS] [EVENT]?} & reference event occurrence & wrong participant binding on a matched event slot \\
Temporal order & \texttt{Did [EVENT 1] happen before [EVENT 2]?} & reference ordered pair & reversed supported order; event collapse/swap conflict \\
\bottomrule
\end{tabular}%
}
\caption{Question templates and supervision sources for the revised verification rewards. The local descriptor $d(o)$ uses the reference phrase for the matched instance and may include a local modifier or index token when it is needed to distinguish repeated instances.}
\label{tab:qa-templates}
\end{table}

\subsubsection{Implementation details of other baselines}
\label{app:other_baseline_details}
For reproduced non-training baselines, we use the same backbone, frame budget, and benchmark protocol as in our main runs and vary only the method-specific decoding or post-hoc scoring knobs. This keeps the comparison focused on the baseline mechanism itself rather than on unrelated prompt or sampling changes.

For TCD~\citep{eventhallusion}, we search a small grid over frame downsampling and contrastive-decoding coefficients. Concretely, we use $r \in \{2,4\}$ and $(\alpha,\beta)\in\{(1.0,0.1),(0.5,0.5)\}$, which gives four configurations in total; for each benchmark, we report the best configuration from this grid. For DINO-HEAL~\citep{vidhalluc}, we search over the two binary choices emphasized by the paper: whether feature normalization is enabled, and whether the DINO backbone uses registers. This again gives four configurations, and we report the best one on each benchmark.

For SEASON~\citep{wu2025season}, we reproduce the method locally for Qwen3-VL and keep the main decoding coefficients fixed across benchmarks. In our runs, we use $n_{\mathrm{frame}}=8$, max\_new\_tokens $=128$, $\alpha=1.0$, $\beta=0.33$, and noise std.\ $=0.05$. The only backbone-specific change is the choice of self-diagnostic attention layers: we use the last four decoder layers of each model, namely 20--23 for Qwen3-VL-4B and Qwen3-VL-8B, and 36--39 for Qwen3-VL-32B. We omit training-based baselines that are evaluated directly from released checkpoints, since their comparison does not depend on a local reimplementation choice.

\subsection{Training Pipeline Analysis}

\subsubsection{Why FACTUAL as the default parser}
\label{app:parser_backend_audit}
We additionally ran a small backend audit to decide whether the reward pipeline should use a local factual parser or an API parser. Because our pipeline parses each caption sentence by sentence before merging sentence-level outputs, we measure parser quality at the sentence level and throughput at the caption level under the same sentence-splitting pipeline. Specifically, we compare the FACTUAL checkpoint~\citep{li2023factual} against GPT-5.4 mini and report the result in Table~\ref{tab:parser_backend_audit}. For quality, we sample 48 training sentences evenly from the six source collections in the TemporalBench short-caption pool and score the resulting graphs with GPT-5.4. Since the factual reward in our final system is driven mainly by object/attribute revisions rather than relation-level graph scoring, the audit focuses on object pass, attribute pass, and an overall atom-pass criterion that asks whether a parse is usable for object/attribute supervision.

Table~\ref{tab:parser_backend_audit} shows that GPT-5.4 mini is the stronger parser on this small audit, especially on attributes, but FACTUAL remains usable for the factual atoms that our reward consumes. On the efficiency side, the local FACTUAL parser is 6.2$\times$ faster than the API parser when measured on 100 training captions with the same sentence-by-sentence pipeline; extrapolated to the full 1{,}891-caption training pool, this reduces a single parsing pass from 109.1 to 17.7 minutes. Using the observed token usage together with the public GPT-5.4 mini pricing in March 2026~\citep{openai_api_pricing_2026}, the same full pass would also require roughly \$3.52 in API cost. We therefore use FACTUAL as the default parser in the training pipeline because it offers the more practical trade-off between parse quality, speed, and operating cost for repeated offline parsing.

\begin{table*}[h]
\centering
\small
\setlength{\tabcolsep}{4.5pt}
\resizebox{\textwidth}{!}{%
\begin{tabular}{lccc|ccc}
\toprule
& \multicolumn{3}{c|}{Sentence-level quality audit (48 sentences)} & \multicolumn{3}{c}{Caption-level efficiency audit (100 captions)} \\
\cmidrule(lr){2-4} \cmidrule(lr){5-7}
Parser & Obj. & Attr. & Atom & Sec./cap. & Full pass (min) & API cost/pass (\$) \\
\midrule
FACTUAL parser & 92.9 & 85.8 & 86.7 & 0.56 & 17.7 & 0.00 \\
GPT-5.4 mini & 89.2 & 91.3 & 87.1 & 3.46 & 109.1 & 3.52 \\
\bottomrule
\end{tabular}%
}
\caption{Sentence-level parser-quality audit and caption-level efficiency audit for choosing the scene-graph parser used by our method. Quality is measured on 48 sampled training sentences from the TemporalBench short-caption pool with a GPT-5.4 judge. \textit{Atom pass} asks whether a parse is usable for the object/attribute factual supervision used by our reward. Efficiency is measured on 100 training captions under the same sentence-by-sentence parsing pipeline and projected to the full 1{,}891-caption training pool. API cost is estimated from observed token usage and the public GPT-5.4 mini pricing in March 2026~\citep{openai_api_pricing_2026}.}
\label{tab:parser_backend_audit}
\end{table*}

\subsubsection{Training-data provenance and overlap audit}
\label{app:data_overlap}
Our default training set is the 1{,}891-example TemporalBench short-caption split, where each caption is paired with a unique video. Concretely, this pool is drawn from the TemporalBench short-caption subset spanning six source collections: ActivityNet, Charades, COIN, EgoExo4D, Movie\_Description, and Oops. In the data-scale ablation, the 150\% setting adds 946 extra captions recovered from the TemporalBench QA splits (70 from `test\_short\_qa' and 876 from `test\_long\_qa'), which introduces 866 additional videos and one extra source collection, FineGym, but still stays inside the TemporalBench video pool.

\begin{table*}[h]
\centering
\scriptsize
\setlength{\tabcolsep}{3.0pt}
\resizebox{\textwidth}{!}{%
\begin{tabular}{lccccccccc}
\toprule
\textbf{Training pool} & \textbf{Caps.} & \textbf{Videos} & \textbf{TempComp.} & \textbf{TVBench} & \textbf{VMME} & \textbf{MVB} & \textbf{VidHall.} & \textbf{EventHall.} & \textbf{Total} \\
\midrule
Default 100\% & 1891 & 1891 & 0/410 & 0/2112 & 0/900 & 0/3851 & 0/1072 & 0/397 & 0/8193 \\
Scaled 150\% & 2837 & 2757 & 0/410 & 0/2112 & 0/900 & 0/3851 & 0/1072 & 0/397 & 0/8193 \\
\bottomrule
\end{tabular}
}
\caption{Train--evaluation overlap audit for the TemporalBench-based training pools. We report exact normalized filename overlap against the official benchmark indices or dataset manifests. `Total' is computed against the union of all six benchmark video sets, which contains 8{,}193 unique videos.}
\label{tab:data_overlap_audit}
\end{table*}

To address the concern that gains may come from benchmark overlap or data leakage, we explicitly audit exact video reuse against the released benchmark indices or official dataset manifests for TempCompass~\citep{tempcompass}, TVBench~\citep{tvbench}, Video-MME~\citep{videomme}, MVBench~\citep{mvbench}, VideoHallucer~\citep{videohallucer}, and EventHallusion~\citep{eventhallusion}. Table~\ref{tab:data_overlap_audit} reports the result. For both the default 100\% training pool and the enlarged 150\% pool, we find zero exact video overlap with all six benchmarks after normalizing video filenames. The union of these benchmark video sets contains 8{,}193 unique evaluation videos, so the observed gains on the audited benchmarks cannot be explained by direct reuse of evaluation clips in training. We likewise use the remaining held-out benchmarks strictly for evaluation and never incorporate their annotations into the training set.

\subsection{Additional Experimental Results}

\subsubsection{Additional VDC Caption-Generation Results}
\label{app:vdc_caption}
Because the motivation of our method originates from video captioning, we additionally evaluate on the VDC caption-generation benchmark using the default Qwen3-VL-8B backbone. Table~\ref{tab:main_vdc_caption} reports the GPT-4o-mini judged score and binary accuracy for five caption types. Our method yields a small but consistent gain over the base model: the average judged score improves from 2.274 to 2.319, and the average binary accuracy improves from 43.97 to 44.82. The gain is consistent across short, main-object, detailed, and camera-focused captions, while the background subset is essentially tied. We view this result as useful supporting evidence that the same structured reward improving grounded QA and hallucination-sensitive evaluation also transfers back to the caption-generation setting that originally motivated the method.

\begin{table}[h]
\centering
\small
\setlength{\tabcolsep}{3.4pt}
\renewcommand{\arraystretch}{0.96}
\begin{tabular}{lcccc}
\toprule
\multirow{2}{*}{\textbf{Caption Type}} & \multicolumn{2}{c}{\textbf{Score} $\uparrow$} & \multicolumn{2}{c}{\textbf{Acc.} $\uparrow$} \\
\cmidrule(lr){2-3} \cmidrule(lr){4-5}
& \textbf{Base} & \textbf{+\methodname} & \textbf{Base} & \textbf{+\methodname} \\
\midrule
Short & 2.274 & 2.348 & 44.04 & 45.49 \\
Main object & 2.345 & 2.404 & 45.50 & 46.51 \\
Detailed & 2.512 & 2.553 & 48.74 & 49.50 \\
Camera & 2.125 & 2.171 & 40.81 & 41.85 \\
Background & 2.116 & 2.117 & 40.74 & 40.73 \\
\midrule
Average & 2.274 & 2.319 & 43.97 & 44.82 \\
\bottomrule
\end{tabular}
\caption{VDC caption-generation results on the default Qwen3-VL-8B backbone. We report the GPT-4o-mini judged mean score (0--5) and binary accuracy for each caption type.}
\label{tab:main_vdc_caption}
\end{table}

\subsubsection{Additional Scaling Results}
\label{app:additional_scaling}
Table~\ref{tab:ablation_data_scale} gives the complete per-benchmark numbers behind the data-scaling curve in Figure~4. The main pattern is that increasing the RL-stage caption pool used by our method improves all three benchmark groups, with the clearest absolute gains on the hallucination-oriented evaluation. The temporal and conventional gains are smaller but remain positive, which is consistent with the view that extra training data mostly helps factual support and temporal grounding rather than generic answer-format adaptation.

Table~\ref{tab:ablation_model_size} reports the full model-scaling results that underlie the backbone-size trend in Figure~4. The 32B backbone is already substantially stronger before RL, so the post-training margin becomes smaller, but the improvement direction remains stable on the available benchmarks. This is the expected scaling behavior for a post-training method that mainly repairs structured consistency rather than changing the underlying visual encoder capacity.

\newcommand{\avgdelta}[2]{$#1_{\scriptstyle #2}$}
\newcommand{\twoline}[2]{\shortstack[c]{\textbf{#1}\\\textbf{#2}}}

\begin{table}[h]
   \centering
   \scriptsize
   \setlength{\tabcolsep}{1.8pt}
   \renewcommand{\arraystretch}{0.92}
   \resizebox{\linewidth}{!}{%
   \begin{tabular}{lccccccccccc}
      \toprule
      & & \multicolumn{3}{c}{\twoline{Temporal}{Understanding}}
        & \multicolumn{3}{c}{\twoline{Conventional Video}{Understanding}}
        & \multicolumn{3}{c}{\twoline{Hallucination}{Examination}} & \\
      \cmidrule(lr){3-5} \cmidrule(lr){6-8} \cmidrule(lr){9-11}
      \textbf{Scale} & \textbf{N}
      & \twoline{Temp}{Compass}
      & \textbf{TVBench}
      & \textbf{Avg.}
      & \twoline{Video-}{MME}
      & \textbf{MVBench}
      & \textbf{Avg.}
      & \twoline{Video}{Hallucer}
      & \twoline{Event}{Hallusion}
      & \textbf{Avg.}
      & \textbf{Avg.} \\
      \midrule
      Baseline & --   & 74.1 & 48.6 & \avgdelta{61.35}{\pm 0.00} & 56.7 & 65.8 & \avgdelta{61.25}{\pm 0.00} & 55.7 & 66.8 & \avgdelta{61.25}{\pm 0.00} & \avgdelta{61.3}{\pm 0.0} \\
      10\%     & 189  & 74.2 & 48.7 & \avgdelta{61.45}{+0.10} & 56.9 & 65.9 & \avgdelta{61.40}{+0.15} & 55.8 & 72.1 & \avgdelta{63.95}{+2.70} & \avgdelta{62.3}{+1.0} \\
      25\%     & 473  & 74.4 & 48.7 & \avgdelta{61.55}{+0.20} & 56.9 & 66.3 & \avgdelta{61.60}{+0.35} & 55.9 & 72.1 & \avgdelta{64.00}{+2.75} & \avgdelta{62.4}{+1.1} \\
      50\%     & 946  & 74.5 & 49.1 & \avgdelta{61.80}{+0.45} & 56.9 & 66.6 & \avgdelta{61.75}{+0.50} & 56.2 & 72.8 & \avgdelta{64.50}{+3.25} & \avgdelta{62.7}{+1.4} \\
      100\%    & 1891 & 75.1 & \textbf{49.2} & \avgdelta{\mathbf{62.15}}{+0.80} & 57.4 & 66.7 & \avgdelta{62.05}{+0.80} & 56.8 & \textbf{73.4} & \avgdelta{\mathbf{65.90}}{+4.65} & \avgdelta{\mathbf{63.4}}{+2.1} \\
      150\%    & 2837 & \textbf{75.2} & 49.1 & \avgdelta{\mathbf{62.15}}{+0.80} & \textbf{57.5} & \textbf{66.8} & \avgdelta{\mathbf{62.15}}{+0.90} & \textbf{57.0} & 75.2 & \avgdelta{66.10}{+4.85} & \avgdelta{63.5}{+2.2} \\
      \bottomrule
   \end{tabular}%
   }
   \caption{Ablation study on training data scale. Each group-wise Avg. is the arithmetic mean of the two benchmarks in that group, and the final Avg. is the arithmetic mean over all six benchmarks. Subscripts indicate the change relative to Baseline.}
   \label{tab:ablation_data_scale}
\end{table}

\begin{table*}[h]
\centering
\footnotesize
\setlength{\tabcolsep}{8pt}
\begin{tabular}{llcccccc}
\toprule
\multirow{2}{*}{\textbf{Category}} & \multirow{2}{*}{\textbf{Benchmark}} & \multicolumn{3}{c}{\textbf{Base}} & \multicolumn{3}{c}{\textbf{Ours}} \\
\cmidrule(lr){3-5}\cmidrule(lr){6-8}
& & \textbf{4B} & \textbf{8B} & \textbf{32B} & \textbf{4B} & \textbf{8B} & \textbf{32B} \\
\midrule
\multirow{2}{*}{\textbf{Temporal}}
& TempCompass    & 71.0 & 74.1 & 77.2 & 73.0$_{\scriptstyle +2.0}$ & 75.1$_{\scriptstyle +1.0}$ & 77.3$_{\scriptstyle +0.1}$ \\
& TVBench        & 41.7 & 47.0 & 51.3 & 47.1$_{\scriptstyle +5.4}$ & 48.0$_{\scriptstyle +1.0}$ & 51.4$_{\scriptstyle +0.1}$ \\
\midrule
\multirow{2}{*}{\textbf{Conventional}}
& Video-MME      & 53.6 & 56.7 & 60.1 & 54.4$_{\scriptstyle +0.8}$ & 57.4$_{\scriptstyle +0.7}$ & 60.3$_{\scriptstyle +0.2}$ \\
& MVBench        & 62.7 & 65.8 & 69.5 & 64.2$_{\scriptstyle +1.5}$ & 66.7$_{\scriptstyle +0.9}$ & 69.8$_{\scriptstyle +0.3}$ \\
\midrule
\multirow{2}{*}{\textbf{Hallucination}}
& VideoHallucer  & 51.4 & 55.7 & 62.0 & 55.8$_{\scriptstyle +4.4}$ & 56.8$_{\scriptstyle +1.1}$ & 62.5$_{\scriptstyle +0.5}$ \\
& EventHallusion & 68.4 & 66.8 & 77.0 & 72.0$_{\scriptstyle +3.6}$ & 73.4$_{\scriptstyle +6.6}$ & 78.0$_{\scriptstyle +1.0}$ \\
\bottomrule
\end{tabular}
\caption{Model-size ablations for Qwen3-VL, grouped into temporal understanding, conventional video understanding, and hallucination-oriented evaluation.}
\label{tab:ablation_model_size}
\end{table*}

\subsubsection{Additional Ablation Tables}
\label{app:additional_ablation_tables}
Table~\ref{tab:ablation_components} provides the numeric version of the reward-component ablation summarized in Figure~3. The main trend matches the grouped plot: the full components remains strongest overall, and each reward branch contributes a different part of the final gain profile.

\begin{table}[h]
   \centering
   \footnotesize
   \setlength{\tabcolsep}{3pt}
   \renewcommand{\arraystretch}{1.03}
   \resizebox{\linewidth}{!}{%
   \begin{tabular}{@{}lccc ccc ccc@{}}
   \toprule
   & \multicolumn{3}{c}{\textbf{Temporal Understanding}} 
   & \multicolumn{3}{c}{\textbf{Conventional Video Understanding}} 
   & \multicolumn{3}{c}{\textbf{Hallucination Examination}} \\
   \cmidrule(lr){2-4} \cmidrule(lr){5-7} \cmidrule(lr){8-10}
   \textbf{Setting} 
   & \textbf{TempCompass} & \textbf{TVBench} & \textbf{Avg}
   & \textbf{VideoMMe} & \textbf{MVBench} & \textbf{Avg}
   & \textbf{VideoHallucer} & \textbf{EventHallusion} & \textbf{Avg} \\
   \midrule
   Ours 
   & 75.1 & 48.0 & 61.55
   & 57.4 & 66.7 & 62.05
   & 56.8 & 73.4 & 65.10 \\
   \midrule
   Qwen3-VL-8B (Base) 
   & 74.1 & 47.0 & 60.55$_{\scriptstyle -1.00}$
   & 56.7 & 65.8 & 61.25$_{\scriptstyle -0.80}$
   & 55.7 & 66.8 & 61.25$_{\scriptstyle -3.85}$ \\
   w/o SG Reward 
   & 74.6 & 47.4 & 61.00$_{\scriptstyle -0.55}$
   & 57.0 & 66.1 & 61.55$_{\scriptstyle -0.50}$
   & 56.2 & 71.3 & 63.75$_{\scriptstyle -1.35}$ \\
   w/o Temp Reward 
   & 74.2 & 47.2 & 60.70$_{\scriptstyle -0.85}$
   & 57.1 & 66.5 & 61.80$_{\scriptstyle -0.25}$
   & 56.6 & 71.5 & 64.05$_{\scriptstyle -1.05}$ \\
   w/o VQA Reward 
   & 74.8 & 47.7 & 61.25$_{\scriptstyle -0.30}$
   & 57.2 & 65.9 & 61.55$_{\scriptstyle -0.50}$
   & 56.0 & 69.9 & 62.95$_{\scriptstyle -2.15}$ \\
   \bottomrule
   \end{tabular}%
   }
   \caption{Ablation studies on our training strategy. SG refers to the scene graph reward. Benchmarks are grouped into temporal understanding, conventional video understanding, and hallucination.}
   \label{tab:ablation_components}
\end{table}

\subsubsection{Additional Qualitative Study}
\label{app:qualitative_study}
\begin{figure*}[h]
    \centering
    \includegraphics[width=\textwidth]{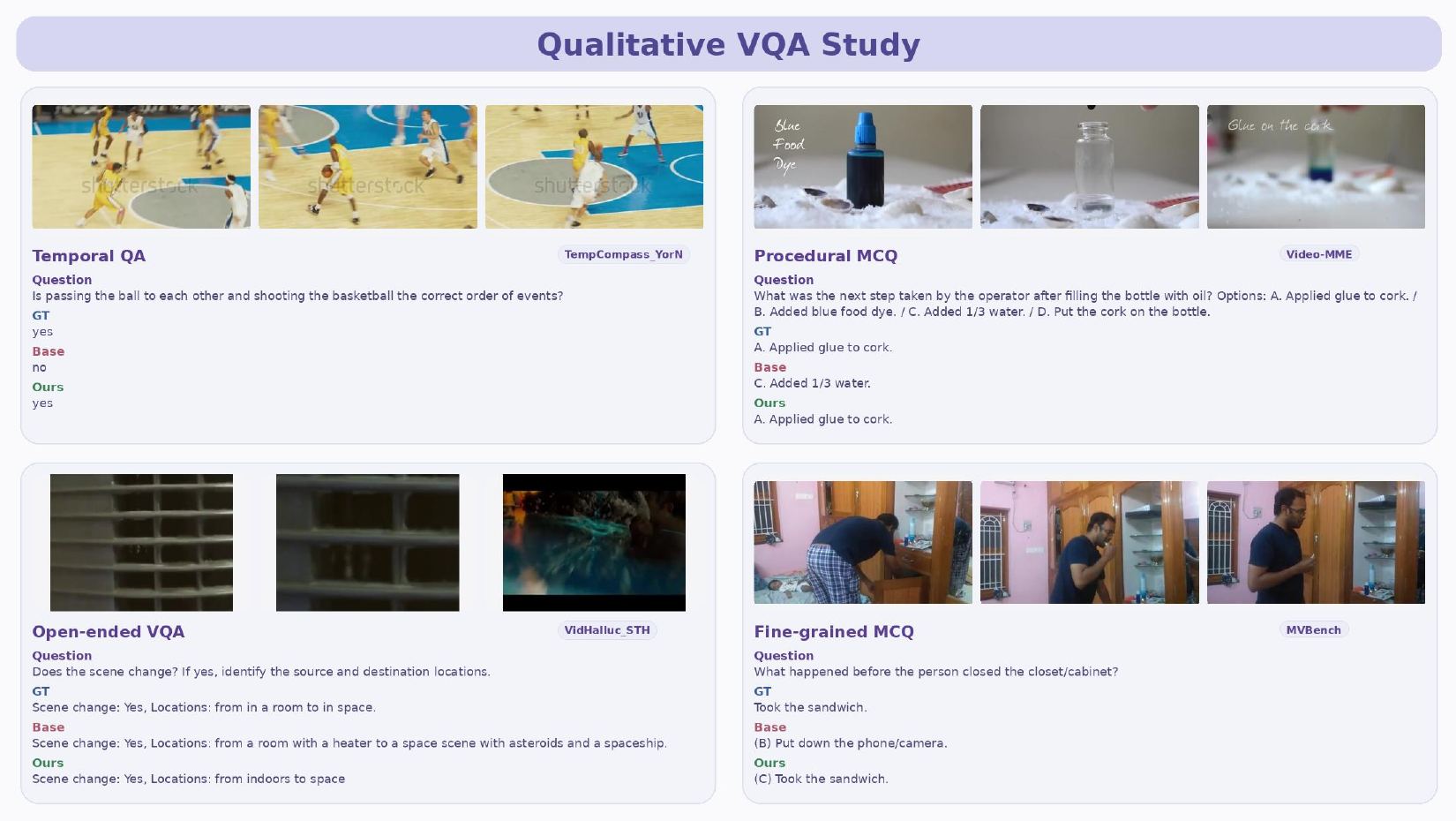}
    \caption{Compact qualitative QA and verification comparison. The cases cover temporal QA, procedural MCQ, fine-grained MCQ, and open-ended scene-change reasoning. Across all four examples, the base model makes a locally plausible but grounded error, whereas our method recovers the correct answer with less ambiguity.}
    \label{fig:appendix_qual_qa}
\end{figure*}

\begin{figure*}[h]
    \centering
    \includegraphics[width=\textwidth]{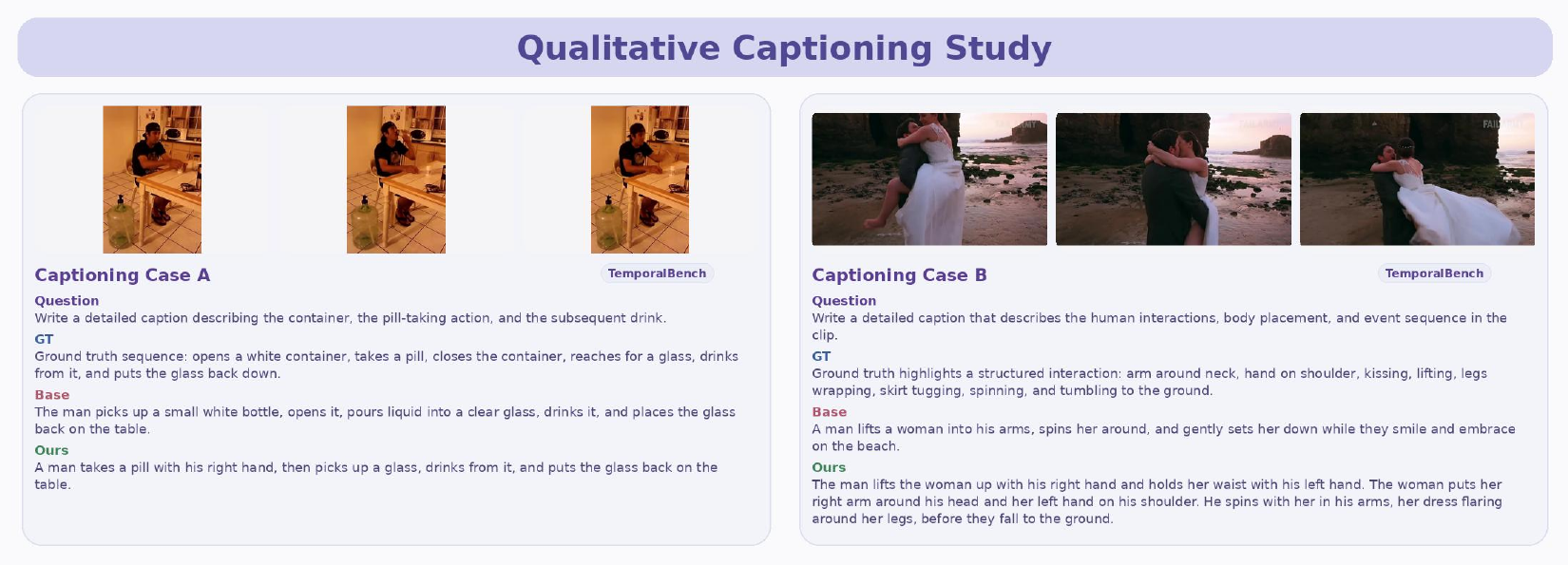}
    \caption{Compact qualitative captioning comparison. The selected cases are drawn from the stored TemporalBench short-caption outputs and highlight failures where the base model produces a plausible but factually wrong long caption, while our method restores the missing action or interaction details.}
    \label{fig:appendix_qual_captioning}
\end{figure*}

To evaluate the results of our method, we provide a compact qualitative comparison based on representative examples drawn from the saved model outputs on benchmark videos. We highlight cases in which the base model fails despite clear visual evidence---for example, on object identity, procedural steps, temporal order, or scene transitions---while our method corrects the same error for the identical video input. Figures~\ref{fig:appendix_qual_captioning} and~\ref{fig:appendix_qual_qa} present two long-form captioning examples and four QA-style examples, respectively.

\end{document}